\documentclass[pdflatex,sn-mathphys-num,iicol]{sn-jnl}

\usepackage{graphicx}%
\usepackage{multirow}%
\usepackage{amsmath,amssymb,amsfonts}%
\usepackage{amsthm}%
\usepackage{mathrsfs}%
\usepackage[title]{appendix}%
\usepackage{xcolor}%
\usepackage{textcomp}%
\usepackage{manyfoot}%
\usepackage{booktabs}%
\usepackage{algorithm}%
\usepackage{algorithmicx}%
\usepackage{algpseudocode}%
\usepackage{listings}%
\usepackage{bm}
\usepackage{subcaption, overpic, textpos}
\usepackage{xspace}

\makeatletter
\DeclareRobustCommand\onedot{\futurelet\@let@token\@onedot}
\def\@onedot{\ifx\@let@token.\else.\null\fi\xspace}

\def\eg{\emph{e.g}\onedot}

\def\wrt{w.r.t\onedot} 
\def\etal{\emph{et al}\onedot}
\makeatother

\newlength\savewidth\newcommand\shline{\noalign{\global\savewidth\arrayrulewidth
  \global\arrayrulewidth 1pt}\hline\noalign{\global\arrayrulewidth\savewidth}}

\renewcommand{\paragraph}[1]{\vspace{1.25mm}\noindent\textbf{#1}}

\usepackage{array}
\newcolumntype{x}[1]{>{\centering\arraybackslash}p{#1pt}}
\newcolumntype{y}[1]{>{\raggedright\arraybackslash}p{#1pt}}
\newcolumntype{z}[1]{>{\raggedleft\arraybackslash}p{#1pt}}

\theoremstyle{thmstyleone}%
\theoremstyle{thmstyletwo}%

\theoremstyle{thmstylethree}%

\begin{document}

\title[Article Title]{AEMIM: Adversarial Examples Meet Masked Image Modeling}


\author[1,2,3]{\fnm{Wenzhao} \sur{Xiang}}\email{xiangwenzhao22@mails.ucas.ac.cn;xiangwzh@pcl.ac.cn}

\author[4]{\fnm{Chang} \sur{Liu}}\email{sunrise6513@sjtu.edu.cn}

\author[5]{\fnm{Hang} \sur{Su}}\email{suhangss@tsinghua.edu.cn}

\author*[2]{\fnm{Hongyang} \sur{Yu}}\email{yuhy01@pcl.ac.cn}

\affil*[1]{\orgdiv{Key Laboratory of Intelligent Information Processing}, \orgname{Institute of Computing Technology (ICT)}, \orgaddress{\street{Chinese Academy of Sciences (CAS)}, \city{Beijing}, \postcode{100190}, \country{China}}}

\affil*[2]{\orgdiv{Peng Cheng Laboratory}, \orgaddress{\city{Shenzhen}, \postcode{10587}, \country{China}}}

\affil[3]{\orgdiv{University of the Chinese Academy of Sciences}, \orgaddress{\city{Beijing}, \postcode{100049}, \country{China}}}

\affil*[4]{\orgdiv{Institute of Image Communication and Networks Engineering}, \orgname{Department of Electronic Engineering}, \orgaddress{Shanghai Jiao Tong University, \city{Shanghai}, \postcode{200240}, \country{China}}}

\affil[5]{\orgdiv{Department of Computer Science and Technology}, \orgname{Tsinghua University}, \orgaddress{ \city{Beijing}, \postcode{100084}, \country{China}}}


\abstract{Masked image modeling (MIM) has gained significant traction for its remarkable prowess in representation learning. As an alternative to the traditional approach, the reconstruction from corrupted images has recently emerged as a promising pretext task. However, the regular corrupted images are generated using generic generators, often lacking relevance to the specific reconstruction task involved in pre-training. Hence, reconstruction from regular corrupted images cannot ensure the difficulty of the pretext task, potentially leading to a performance decline. Moreover, generating corrupted images might introduce an extra generator, resulting in a notable computational burden. To address these issues, we propose to incorporate adversarial examples into masked image modeling, as the new reconstruction targets. Adversarial examples, generated online using only the trained models, can directly aim to disrupt tasks associated with pre-training. Therefore, the incorporation not only elevates the level of challenge in reconstruction but also enhances efficiency, contributing to the acquisition of superior representations by the model. In particular, we introduce a novel auxiliary pretext task that reconstructs the adversarial examples corresponding to the original images. We also devise an innovative adversarial attack to craft more suitable adversarial examples for MIM pre-training. It is noted that our method is not restricted to specific model architectures and MIM strategies, rendering it an adaptable plug-in capable of enhancing all MIM methods. Experimental findings substantiate the remarkable capability of our approach in amplifying the generalization and robustness of existing MIM methods. Notably, our method surpasses the performance of baselines on various tasks, including ImageNet, its variants, and other downstream tasks.}




\maketitle

\section{Introduction}
Self-supervised learning (SSL) has achieved phenomenal success in natural language processing (NLP) tasks~\cite{radford2018improving,radford2019language,brown2020language,kenton2019bert}. Inspired by masked language modeling, masked image modeling (MIM) has received a great deal of attention in the computer vision community~\cite{bao2021beit,he2022masked,xie2022simmim}. Similarly, MIM masks a portion of image patches and then reconstructs the masked patches of the original data via an auto-encoder architecture. And the models can learn rich visual representations in this way. Compared with contrastive learning methods~\cite{oord2018representation,he2020momentum,chen2020simple}, SSL methods based on masked image modeling exhibit enhanced performance across a wide range of downstream vision tasks.

Several recent studies~\cite{fang2023corrupted, Brempong_2022_CVPR, you2023beyond} propose adopting the reconstruction of corrupted images as a pretext task for pre-training, as an alternative to the traditional approach of reconstructing masked normal images. Their findings indicate that reconstructing corrupted images can assist the model in acquiring more generalized representations. However, the process of generating corrupted images might necessitate an auxiliary generator~\cite{fang2023corrupted}, thereby imposing a significant extra computational burden. 
Moreover, these regular corrupted images are created using generic generators, frequently without direct relevance to the specific reconstruction task of the pre-training process. Compared to reconstructing masked normal images, reconstructing regular corrupted images does not inherently increase the difficulty of the pretext task, but rather transforms the distribution of reconstructed images. Therefore, they cannot serve as consistently challenging reconstruction targets. This might result in some performance loss, as engaging with more challenging pretext tasks can significantly facilitate the model in acquiring superior representations~\cite{wang2023hard}.

A straightforward approach to synthesize the more challenging targets is to leverage adversarial examples~\cite{goodfellow2014explaining,madry2018towards,dong2018boosting}. Adversarial examples are deliberately crafted to lower a model's performance on particular tasks. Their existence undeniably introduces substantial hurdles to task fulfillment. These adversarial examples are widely acknowledged as hard negative samples, rendering them more intricate reconstruction targets for pre-training. Additionally, adversarial examples are often generated online, utilizing only the model currently being trained and eliminating the need for an additional generator, thereby significantly enhancing training efficiency.

\begin{figure*}
\begin{center}
\includegraphics[width=\linewidth]{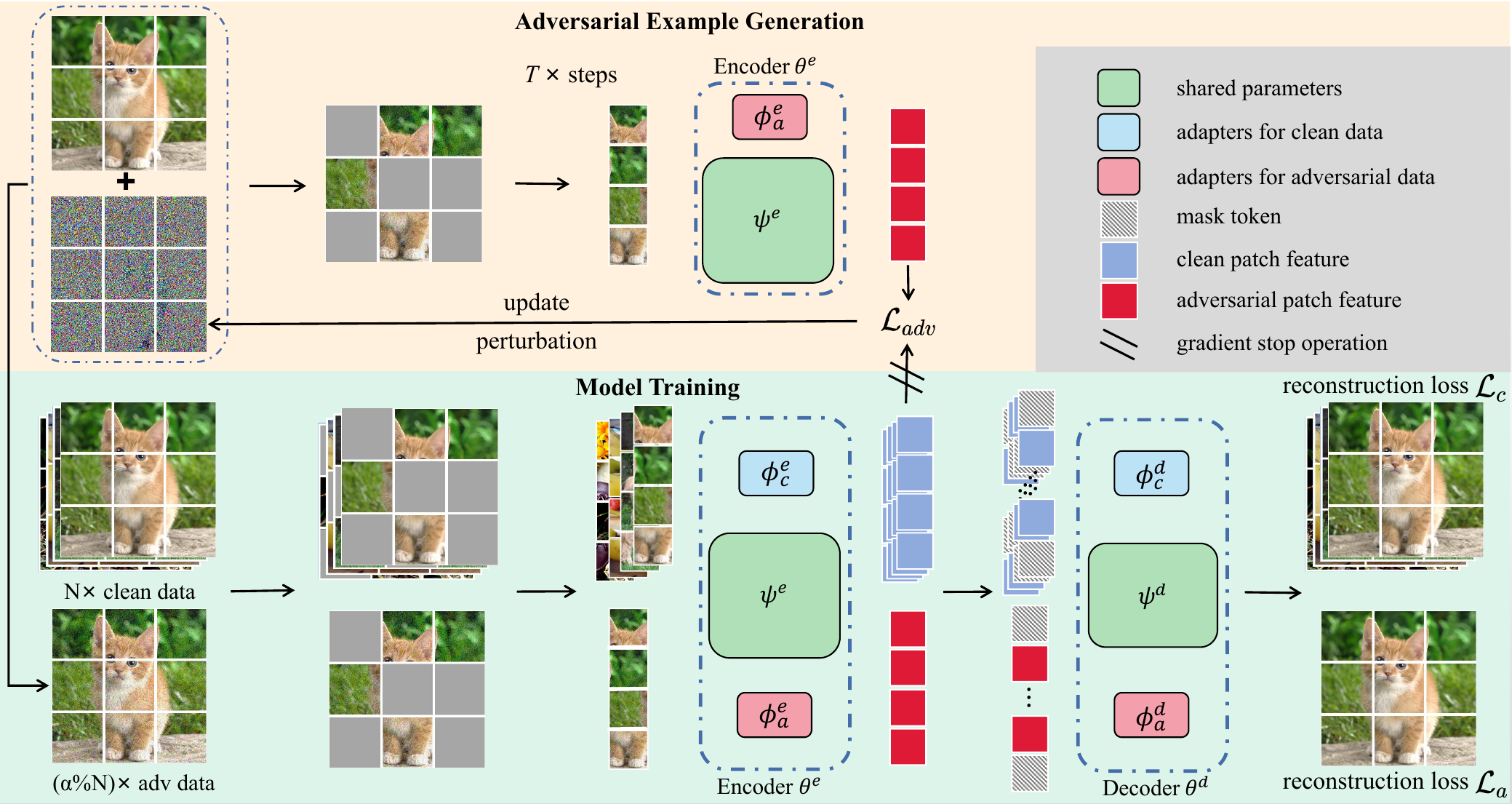}
\end{center}
\caption{Overall pipeline of the proposed AEMIM method. Our method extends the normal MIM method (taking MAE as an example) by constructing an auxiliary pretext task of reconstructing images from their corresponding adversarial examples. And adapters are introduced to handle clean data and adversarial data, respectively. Our method consists of two parts: 1) Adversarial example generation. The adversarial examples are generated using the distance between the encoder's feature representation of the adversarial data and that of the clean data as the adversarial loss. 2) Model training. The model is trained with two reconstruction pretext tasks for clean data and adversarial data. The shared parameters $\boldsymbol{\psi^{e}}$ and clean adapters $\boldsymbol{\phi^{e}_c}$ of the encoder are the pre-trained parameters that are ultimately retained for fine-tuning on downstream tasks.}
\label{fig:1}
\end{figure*}

Nevertheless, studies~\cite{madry2018towards, zhang2019theoretically, liu2023comprehensive} on adversarial training have revealed that if adversarial examples are directly used for training, the performance of the model on normal vision tasks will significantly decline, even though its robustness could be enhanced. Fortunately, some recent studies~\cite{xie2020adversarial,rebuffi2023revisiting}, like AdvProp~\cite{xie2020adversarial} have explored the advantages of utilizing adversarial examples to enhance normal visual tasks. These studies demonstrate that co-training with both clean and adversarial images, and utilizing a small set of independent model parameters (known as "adapters") to handle inputs from different domains, can improve the performance of models on various visual tasks~\cite{xie2020adversarial,mei2022fast,chen2021robust}. 
The researchers view adversarial examples as a regularizer that can prevent over-fitting during training. We believe that utilizing adversarial examples as the reconstruction targets in conjunction with adapters can function as a regularization technique for MIM pre-training, ultimately leading to improved representation learning.

In this paper, we propose to integrate adversarial examples into masked image modeling and explore their potential benefits in representation learning. We create an auxiliary pretext task of reconstructing corrupted images to provide regularization for the original pretext task and adopt the more challenging adversarial examples as the corrupted images. Adapters are employed in the model to handle data from different source domains.
We develop a novel adversarial attack for MIM pre-training that can synthesize adversarial examples without relying on ground-truth labels, which are not available in SSL.
To elaborate further, we employ the distance between the encoder's feature representation of adversarial and clean data as the adversarial loss for generating adversarial examples. This is based on our empirical observation that directly employing the reconstruction loss as adversarial loss leads the model to prioritize learning the auxiliary pretext task during pre-training, consequently impeding the learning of the primary pretext task.
To train our model, we follow the AdvProp framework and adopt normalization layers as adapters to capture distinct adversarial and clean data statistics. 
Our proposed method, which boosts Masked Image Modeling with Adversarial Examples (denoted by the prefix "AE-", \eg "AEMIM") as illustrated in Figure~\ref{fig:1}, is not constrained to specific model architectures or MIM strategies, making it a flexible plug-in that can effectively enhance the performance of various MIM methods across different model architectures. We also offer an expedited version of our algorithm to improve training efficiency while retaining most of the performance.

To verify the effectiveness of our method, we conduct experiments on the ImageNet-1K~\cite{russakovsky2015imagenet}, ImageNet independent and identically 
distributed~(IID) variants (ImageNet-Real~\cite{beyer2020we}, ImageNet-V2~\cite{recht2019imagenet}), ImageNet Out-of-Distribution~(OOD) variants (ImageNet-A~\cite{hendrycks2021natural}, ImageNet-R~\cite{hendrycks2021many}, ImageNet-Sketch~\cite{wang2019learning}, ImageNet-C~\cite{hendrycks2018benchmarking}) and COCO~\cite{lin2014microsoft} datasets. 
In our experiments, we adopt MAE as the basic MIM framework for enhancement and simply denote it as AEMIM unless otherwise specified. 
Our AEMIM achieves 83.8\% Top-1 accuracy on ImageNet-1K with ViT-B/16 model at 800 epochs, outperforming the baseline MAE by 0.5/0.2\% at 800/1600 epochs. For ImageNet IID and OOD variants, our best ViT-B/16 model outperforms MAE by an average of 0.6\% and 2.2\%. Similar outcomes can be observed in other downstream tasks. The fast version of AEMIM also achieves competitive results with much higher efficiency.

Our contributions are summarized as follows:
\begin{itemize}\setlength{\itemsep}{-1pt}
    \item Our proposed method, AEMIM, offers a novel framework for improving MIM pre-training, by introducing an auxiliary pretext task of reconstructing masked adversarial examples or their features. Our approach can be used as a plug-in to enhance the performance of existing MIM methods. To our knowledge, this is the first attempt to improve MIM pre-training with adversarial examples for vision tasks. We also provide a faster version of our method for efficiency.
    \item We introduce a novel method for generating adversarial examples specifically tailored for MIM pre-training. Our approach utilizes the distance between clean and adversarial data in the encoder's feature representation as the adversarial loss, eliminating the need for ground-truth labels.
    \item We conduct comprehensive experiments to show that our proposed method can assist the pre-trained models in acquiring more generalized and robust representations. Our method outperforms the baseline on various datasets.
\end{itemize}

\section{Background}
\subsection{Masked Image Modeling}

Inspired by the significant success of masked language modeling (MLM) in NLP like BERT~\cite{kenton2019bert}, masked image modeling (MIM) similarly learns rich vision representations by predicting the raw information from the masked data. BEiT~\cite{bao2021beit} introduces an offline tokenizer VQ-VAE~\cite{van2017neural} to tokenize the image into discrete vision tokens first and set up a patch-level dictionary like in MLM. The pretext task is to recover the vision token ids of the masked patches. PeCo~\cite{dong2023peco} proposes to train the visual tokenizer with a perceptual similarity regularization term to learn a perceptual codebook for pre-training. Instead of using an offline tokenizer, iBOT~\cite{zhou2022image} set up an online tokenizer to produce the target for the encoder, building a self-distillation task for pre-training.

Another series of work directly predicts the raw pixel values or auxiliary features of the masked image patches. MAE~\cite{he2022masked} and SimMIM~\cite{xie2022simmim} propose to reconstruct the raw pixel values from only visible patches and all patches, respectively. CAE involves an encoder-regressor-decoder architecture, where the regressor predicts the representations of masked patches. Instead of predicting raw pixel values directly, MaskFeat~\cite{wei2022masked} propose to use low-level features as the reconstruction target, \eg Histograms of Oriented Gradients (HOG)~\cite{dalal2005histograms}. Researches further explore applying high-level features as the target, such as pre-trained DINO~\cite{gao2023mimic}, momentum features~\cite{baevski2022data2vec}, and multi-modality features~\cite{wei2022mvp,hou2022milan,gao2023mimic}. Unlike other MIM methods, AEMIM introduces an auxiliary proxy task to reconstruct adversarial examples or their high-level representations. Our framework is practically independent of specific MIM strategies, allowing it to be seamlessly integrated as a plugin into almost any MIM method.

There are also some works that integrate the ideas of adversarial learning. ADIOS~\cite{shi2022adversarial} introduces an extra occlusion model to mask the images in an adversarial manner for MIM. NIM~\cite{you2023beyond} adopts denoising as the pretext task and reconstructs noisy images to improve the adversarial robustness.
CIM~\cite{fang2023corrupted} is an effective MIM method that involves the reconstruction of corrupted images. It employs a trainable generator and a pre-trained frozen tokenizer to initially corrupt the input images and then performs the pre-training by recovering all the original image pixels, or predicting whether each visual token is corrupted or not. The generator and the backbone model are trained simultaneously and updated synergistically, resulting in a significant computational burden and making the training process more difficult to converge to the optimal point. Compared to the above methods, AEMIM does not require an additional generator to produce adversarial examples. Instead, it generates them online using adversarial loss based on the training model, making AEMIM both simpler and more efficient.

Most of the current MIM methods utilize an auto-encoder architecture. Let $\mathcal{F}$ denote the auto-encoder parameterized by $\boldsymbol{\theta}$, and $\mathcal{L}_{mim}$ denote the reconstruction loss. For example, in the case of reconstructing raw pixel values, the objective function for MIM pre-training is written as
\begin{equation}
\label{eq:3}
\mathop{\min} \limits_{\boldsymbol{\theta}} \mathbb{E}_{\boldsymbol{x} \sim \mathbb{D}} \mathcal{L}_{mim}(\mathcal{F}(\boldsymbol{x}^m;\boldsymbol{\theta}), \boldsymbol{x}),
\end{equation}
where $\mathbb{D}$ is the data distribution, and $(\cdot)^m$ denotes the masking operation. 

\subsection{Adversarial Learning and Vision Tasks}
\label{sec:2.2}

Adversarial examples can greatly degrade the performance of models by adding imperceptible perturbations to clean images. The goal of generating an adversarial example $\boldsymbol{x}_{a}$ is to find the one that can fool the model to predict the wrong result but is imperceptible to humans, which is expressed as 
\begin{equation}
\label{eq:1}
\mathop{\max} \limits_{\boldsymbol{x}_{a} \in \mathbb{S}} \mathcal{L}_{adv}(\boldsymbol{x}_{a}, \boldsymbol{y},\boldsymbol{\theta}_f), 
\end{equation}
where $\boldsymbol{x}, \boldsymbol{y}, \boldsymbol{x}_{a}$ denote the original image, the ground-truth label and the adversarial example, $\mathcal{L}_{adv}$ denote the adversarial loss function,   $\boldsymbol{\theta}_f$ denote the parameters of the model, $\mathbb{S}=\{\boldsymbol{x}_{a} \ | \ \|\boldsymbol{x}_{a}-\boldsymbol{x}\|_p \leq \epsilon \}$, and $\epsilon$ is the perturbation budget.

Under the white-box setting, where the model architecture, parameters, and gradients are accessible, the perturbations are generally generated with the guidance of gradients. FGSM~\cite{goodfellow2014explaining} is the most famous one-step adversarial attack to generate adversarial examples simply and fast. PGD~\cite{madry2018towards} extend FGSM to an iterative version, further improving the attack success rate of adversarial examples. C\&W~\cite{carlini2017towards} formulates the problem in Lagrangian form and adopts Adam for optimization. There are also many black-box attacks, such as transfer-based attacks~\cite{dong2018boosting,xie2019improving,dong2019evading,Lin2020Nesterov,xiang2023improving}, which generate the adversarial examples with a known source model and then use them to attack the unknown target model.
Recent research is also exploring adversarial examples with high image quality that are closer to real-world data distributions~\cite{kazemi2023minimally,liu2023improving}.
Among them, PGD is the most commonly used method for adversarial training. PGD simply performs the iterative updates as 
\begin{equation}
\label{eq:2}
    \boldsymbol{x}_{a}^{t+1} = \Pi_{\mathcal{B}_p({x}, \epsilon)} \left(\boldsymbol{x}_{a}^{t} + \mu \cdot  \mathrm{sign}\left(\nabla_x \mathcal{L}_{adv}(\boldsymbol{x}_{a}^{t}, \boldsymbol{y},\boldsymbol{\theta}_f)\right)\right), 
\end{equation}
where $\nabla_x\mathcal{L}_{adv}(\cdot, \cdot,\cdot)$ is the gradient of the loss function $\mathcal{L}_{adv}$ \wrt $\boldsymbol{x}$, $\mathrm{sign}(\cdot)$ is the sign function, $\Pi$ is the projection operation, $\mathcal{B}_p(\boldsymbol{x}, \epsilon)$ is the $L_p$ ball centered at $\boldsymbol{x}$ with radius $\epsilon$, and $\mu$ is the step size.

To facilitate the introduction of our approach, we revisit the co-training of clean and adversarial images in the classification task, which is proposed in FGSM. The objective function is expressed as
\begin{align}
\label{eq:4}
\mathop{\min} \limits_{\boldsymbol{\theta}} \mathbb{E}_{(\boldsymbol{x},\boldsymbol{y}) \sim \mathbb{D}} [ &\lambda \mathcal{L}_{cls}(\boldsymbol{f}(\boldsymbol{x};\boldsymbol{\theta}_f), \boldsymbol{y})+ \\ \nonumber 
&(1-\lambda) \mathop{\max} \limits_{\boldsymbol{x}_{a} \in \mathbb{S}} \mathcal{L}_{cls}(\boldsymbol{f}(\boldsymbol{x}_{a};\boldsymbol{\theta}_f), \boldsymbol{y}) ],
\end{align}
where $\boldsymbol{f}(\cdot;\boldsymbol{\theta}_f)$ is a classifier parameterized by $\boldsymbol{\theta}_f$, $\mathcal{L}_{cls}$ is the classification loss, $\lambda$ is the loss ratio, and $\mathbb{S}$ defines the legitimate range of adversarial perturbations as in Equation~\ref{eq:1}. Due to the absence of ground truth in SSL, our AEMIM method proposes using the distance between the representations of clean and adversarial data as the adversarial loss.

Normal adversarial training can significantly improve the adversarial robustness of models. However, the performance of models on normal vision tasks often experiences a notable decline after being trained with adversarial examples. To address the issue, Xie~\etal propose AdvProp~\cite{xie2020adversarial} framework, which performs co-training with clean and adversarial images together and uses distinct batch normalization layers for clean and adversarial images to accumulate different statistics. They regard the adversarial item as a regularizer. With the AdvProp framework, the performance of models on image recognition task is further enhanced. 
Chen~\etal~\cite{chen2021robust} further adopt AdvProp to object detection, and improves detectors' robustness against distortions and domain shifts. Mei~\etal~\cite{mei2022fast} propose Fast-AdvProp, which revamps the costly training components in AdvProp. Fast-AdvProp greatly accelerates the training process of AdvProp and almost no performance is lost.
Rebuffi~\etal~\cite{rebuffi2023revisiting} integrates adapters to adversarial training. 
They point out that the key element to making the co-training work is to "\textit{have trainable parameters which are specific to the clean and adversarial images}". They show that other types of adapters also fit the co-training in addition to the batch normalization layer, \eg classification token in the vision transformer. 
AEMIM integrates adapters into the co-training of clean data and specially designed adversarial data, further enhancing the generalization and robustness of the pre-trained model.

\section{Methodology}

\subsection{Problem Formulation}
\label{sec:3.1}
Recent research~\cite{fang2023corrupted,Brempong_2022_CVPR,you2023beyond} has highlighted the potential of using the reconstruction of corrupted images as a promising pretext task for pre-training.
We argue that adversarial examples pose more challenging reconstruction targets compared to typical corrupted images, and their generation does not require any additional generator. To retain the benefits of the original pretext task, which involves reconstructing masked normal images, we introduce a novel auxiliary pretext task of reconstructing their corresponding adversarial examples. We also incorporate adapters during pre-training to prevent adversarial examples from compromising the model's performance on normal visual tasks. 

We transfer the objective function of the co-training in Equation~\ref{eq:4} from classification to MIM, where ground-truth labels are unavailable, and introduce adapters into the formulation. In the context of adapter literature, most of the model parameters are shared across data from different domains. Nonetheless, a distinct subset of parameters, known as adapters, exists specifically for data from different domains. In our case, we denote the shared parameters as $\boldsymbol{\psi}$, the adapters for clean images as $\boldsymbol{\phi}_{c}$ and the adapters for adversarial images as $\boldsymbol{\phi}_{a}$. Obviously, the model parameters $\boldsymbol{\theta}$ consist of $\boldsymbol{\psi}$, $\boldsymbol{\phi}_{c}$ and $\boldsymbol{\phi}_{a}$. And we only consider the normalization layer and classification token in the vision transformer as the adapters in our method. Then the objective function of the co-training for MIM is rewritten as
\begin{align}
\label{eq:5}
\mathop{\min} \limits_{\boldsymbol{\theta}} \mathbb{E}_{\boldsymbol{x} \sim \mathbb{D}} [ &\lambda \mathcal{L}_{mim}(\mathcal{F}(\boldsymbol{x}^m;\boldsymbol{\psi}\cup\boldsymbol{\phi}_{c} ), \boldsymbol{x})+ \\ \nonumber
&(1-\lambda) \mathcal{L}_{mim}(\mathcal{F}(\boldsymbol{x}_{a}^m;\boldsymbol{\psi}\cup\boldsymbol{\phi}_{a}), \boldsymbol{x}) ],
\end{align}
where the adversarial example $\boldsymbol{x}_{a}$ is generated online with the target of maximizing adversarial loss $\mathcal{L}_{adv}$. And we provide more details on the design of $\mathcal{L}_{adv}$ in the next section. 

\begin{figure}[h]
\begin{center}
\includegraphics[width=\linewidth]{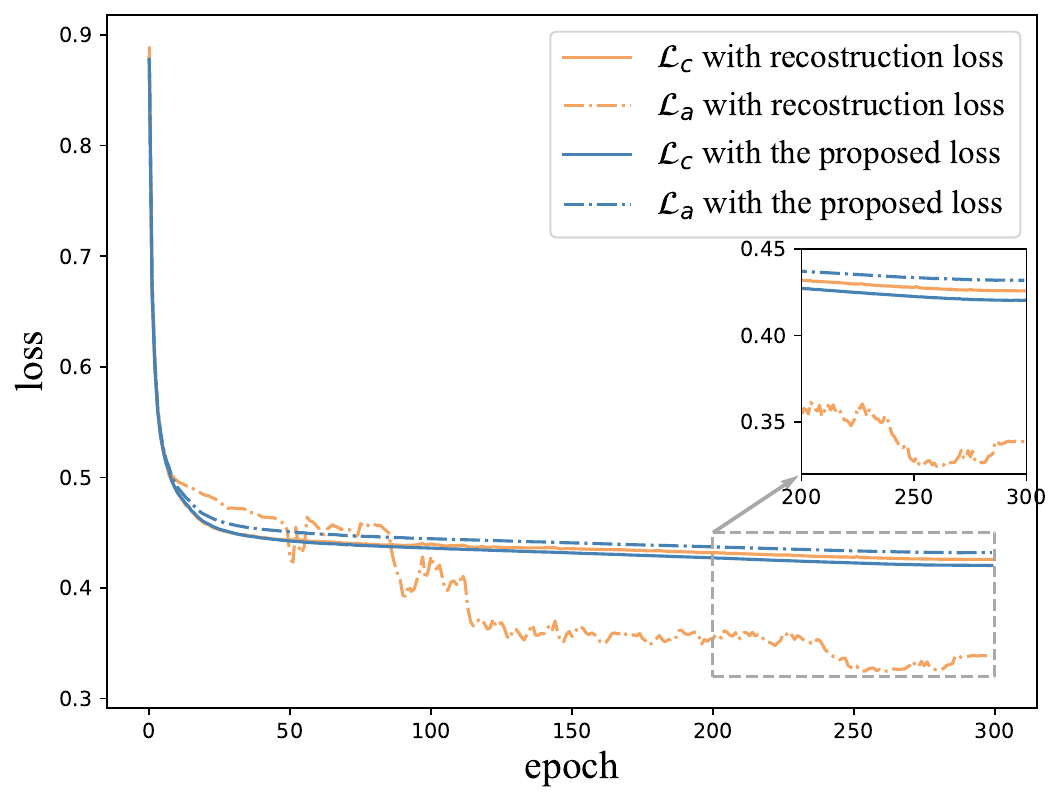}
\end{center}
\caption{Loss curve during pre-training. We illustrate the changes in the loss for the original and auxiliary pretext tasks when utilizing reconstruction loss or our proposed loss as the adversarial loss. }
\label{fig:2}
\end{figure}

Unlike in supervised learning, there are no ground truth labels to guide the generation of adversarial examples during MIM pre-training.
A straightforward approach is to use the task loss directly as the adversarial loss for generating adversarial examples. 
However, our empirical observations suggest that employing the reconstruction loss as the adversarial loss directly can hinder the learning of the original pretext task due to interference from the auxiliary task.
Figure~\ref{fig:2} illustrates the change in loss during the pre-training process. As shown in the figure, the reconstruction loss of adversarial data $\mathcal{L}_{a}$ becomes much lower than that of clean data $\mathcal{L}_{c}$ during training. 
This phenomenon suggests that the model might prioritize solving the auxiliary pretext task during pre-training, which adversely affects the learning of the original pretext task. We believe this is because the aforementioned adversarial examples directly target the original pretext task, resulting in overly strong regularization.

\subsection{Adversarial Examples for Masked Image Modeling}
\label{sec:3.2}

We then design a new adversarial loss, which is only corresponding to the encoder in the MIM framework. We propose to adopt the distance of the encoder's feature representation between adversarial and clean data as the adversarial loss, since: 1) It does not directly target the reconstruction task. 2) The encoder is the backbone model that is finally transferred to downstream tasks as the feature extractor. The adversarial data only need to regularize the encoder to learn generalized representations. 3) Dropping the decoder can accelerate the generation of adversarial examples. 
Denote the encoder as $E$ and the decoder as $D$. The objective function of generating adversarial examples is expressed as 
\begin{gather}
\label{eq:6}
\mathop{\max} \limits_{\boldsymbol{x}_{a} \in \mathbb{S}} \mathcal{L}_{adv}(\boldsymbol{x}, \boldsymbol{x}_{a},\boldsymbol{\theta}^{e}), \\
\label{eq:7}
\mathcal{L}_{adv} = \mathcal{L}_{d}({E}(\boldsymbol{x}_{a}^m,\boldsymbol{\psi}^{e}\cup\boldsymbol{\phi}_{a}^{e}), sg({E}(\boldsymbol{x}^m,\boldsymbol{\psi}^{e}\cup\boldsymbol{\phi}_{c}^{e}))),
\end{gather}
where $\boldsymbol{\theta}^{e}$ is the parameters of the encoder, consisting of the shared parameters $\boldsymbol{\psi}^{e}$, the clean adapters $\boldsymbol{\phi}^{e}_{c}$ and adversarial adapters $\boldsymbol{\phi}^{e}_{a}$, $\mathcal{L}_{d}$ is one type of distance loss, $sg(\cdot)$  means stop-gradient and $\mathbb{S}$ is set to $\mathbb{S}=\{\boldsymbol{x}_{a} \ | \ \|\boldsymbol{x}_{a}-\boldsymbol{x}\|_\infty \leq \epsilon \}$. For $\mathcal{L}_{d}$, we mainly consider $L_2$ distance and $KL$ divergence.
In our method, we apply PGD-$K$ to generate adversarial examples, where $K$ is the iteration number.  
We initialize $\boldsymbol{x}_{a}$ by adding random noise to $\boldsymbol{x}$ in the first step since there is no difference between $\boldsymbol{x}$ and $\boldsymbol{x}_{a}$ at first.
The algorithmic details are presented in Algorithm~\ref{alg:A}.

\subsection{Algorithm Details}
\label{sec:3.3}
\begin{algorithm}[tp]
\caption{AEMIM in Pre-training}
\label{alg:A}
\begin{algorithmic}[1]
\renewcommand{\algorithmicrequire}{\textbf{Input:}}
\renewcommand{\algorithmicensure}{\textbf{Output:}}
\Require Dataset $\mathbb{D}$; an auto-encoder $\mathcal{F}$ with parameters $\boldsymbol{\theta}$, including an encoder ${E}$ with parameters $\boldsymbol{\theta}^{e}$; shared parameters $\boldsymbol{\psi}$ of $\mathcal{F}$; adapters $\boldsymbol{\phi}_c$ of $\mathcal{F}$ for clean images; adapters $\boldsymbol{\phi}_a$ of $\mathcal{F}$ for adversarial images; reconstruction loss $\mathcal{L}_{mim}$; adversarial loss $\mathcal{L}_{adv}$; ratio of adversarial examples $\alpha$; loss ratio $\lambda$; perturbation budget $\epsilon$; number of attack steps $T$; attack step size $\mu$.
\Ensure Network parameter $\boldsymbol{\theta}$.
\For{a batch data of $\boldsymbol{x} \in \mathbb{D}$}
\State{Randomly select $\alpha\%$ of the data: $\boldsymbol{x'}$}
\State{Initializtion with random noise $\delta_x$: $\boldsymbol{x}^0_{a} = \boldsymbol{x'} + \delta_x$
}
\For{$t=0$ to ($T-1$)}
\State{Compute adversarial loss $\mathcal{L}_{adv}(\boldsymbol{x}, \boldsymbol{x}_{a},\boldsymbol{\theta}^{e})$
}
\State{Update adversarial examples with
\begin{align}
    \boldsymbol{x}_{a}^{t+1} = &\Pi_{\mathcal{B}_\infty({x}, \epsilon)} \big\{\boldsymbol{x}_{a}^{t} + \nonumber \\
    &  \mu \cdot  \mathrm{sign}\left(\nabla_x \mathcal{L}_{adv}(\boldsymbol{x}, \boldsymbol{x}_{a},\boldsymbol{\theta}^{e}) \right) \big\} \nonumber
\end{align}
}
\EndFor
\State{Get final adversarial examples: $\boldsymbol{x}_{a} = \boldsymbol{x}^T_{a}$}
\State{Compute loss $\mathcal{L}_{c}=\mathcal{L}_{mim}(\mathcal{F}(\boldsymbol{x}^m;\boldsymbol{\psi}\cup\boldsymbol{\phi}_{c} ), \boldsymbol{x})$ on clean mini-batch data with adapters $\boldsymbol{\phi}_c$}
\State{Compute loss $\mathcal{L}_{a}=\mathcal{L}_{mim}(\mathcal{F}(\boldsymbol{x}_{a}^m;\boldsymbol{\psi}\cup\boldsymbol{\phi}_{a}), \boldsymbol{x})$ on adversarial mini-batch data with adapters $\boldsymbol{\phi}_a$}
\State{Minimize the total loss \wrt network parameters $\boldsymbol{\theta}$
\begin{center}
    $\mathop{\min} \limits_{\boldsymbol{\theta}} \lambda \mathcal{L}_{c} + (1-\lambda)\mathcal{L}_{a}$
\end{center}
}
\EndFor

\end{algorithmic}
\end{algorithm}

In this section, we present the complete process of using adversarial examples to enhance MIM pre-training. Similar to adversarial training in PGD~\cite{madry2018towards}, we formulate the co-training as a min-max game. Each iteration consists of two steps. During the inner maximization step, the model parameters $\boldsymbol{\theta}$ are fixed, while the adversarial examples are generated using the method outlined earlier. During the outer minimization step, both clean images and adversarial images are fed into the model to optimize the parameters $\boldsymbol{\theta}$ through two pretext tasks. 
We also provide an accelerated version of our algorithm. In the stand version, we generate an adversarial image for each clean one, and then perform joint training with the paired data. We also employ $K>1$ for PGD to enhance the quality of adversarial examples. In the expedited version, we only select a portion of the clean data to generate corresponding adversarial examples and set $K=1$ for the PGD attacker, which implies the PGD attacker degrades into an FGSM attacker with random noise initialization. 
In our framework, we ultimately randomly select $\alpha\% = 25\%$ of the clean images to generate corresponding adversarial images. Note the ratio $\alpha\%$ is $100\%$ in the standard version.
We present the above process in the pseudo-code form in Algorithm~\ref{alg:A}, and illustrate the complete pipeline of our proposed framework in Figure~\ref{fig:1}.

\section{Experiments}
\label{sec:4}

\subsection{Experiments Setup}
\label{sec:4.1}
We conduct the pre-training on the training dataset of ImageNet-1K~\cite{russakovsky2015imagenet} without labels. We adopt MAE~\cite{he2022masked} as our basic framework and enhance its performance through our proposed method. Therefore, MAE serves as the main method for comparison in our experiments. It should be emphasized that our method is not confined to particular model architecture or MIM strategies. As such, it can be easily integrated into other MIM methods as a plug-in to enhance their performance. The vanilla ViT-S/16 and ViT-B/16~\cite{dosovitskiy2021an} are utilized as the backbone models for the encoder. After the pre-training stage, we discard all the decoders and only keep the encoders for fine-tuning on downstream tasks. The adapters for adversarial examples are omitted as well, as they could potentially impact the model's clean accuracy. All models are trained on 8 V100 32GB GPUs.

\noindent \textbf{Pre-training setup} We follow most of the MIM settings in MAE. By default, we randomly mask 75\% of the image patches. The decoder consists of 8 transformer layers with 512 feature dimensions and 12 attention heads. We simply use random cropping and horizontal flip as the data augmentations. The models are pre-trained for 300/800 epochs with 15/40 warmup epochs, respectively. And the total batch size is 4096. We adopt AdamW~\cite{loshchilov2018decoupled} optimizer with a base learning rate of $1.5 \times 10^{-4}$, a weight decay of $0.05$, $\beta_1=0.9, \beta_2=0.95$ and cosine learning rate schedule~\cite{loshchilov2017sgdr}. For attack settings, the perturbation budget is set as $\epsilon = 2$ and the number of attack steps is set as $T = 2/1$ for stand AEMIM and Fast AEMIM. We adopt the normalization layer as the default type of adapter and $L_2$ distance as the default distance loss for adversarial examples. And the loss ratio $\lambda$ is empirically set to be 0.5 and the adversarial ratio is set as 100\%/25\% for AEMIM/Fast AEMIM. We simply use random cropping and horizontal flip as the data augmentations. And we use Xavier uniform~\cite{glorot2010understanding} to initialize all Transformer blocks, following MAE. All the default setting is shown in Table~\ref{tab:impl_pretraining}.
\begin{table}[!hbtp]
\centering
\caption{{Pre-training setting on ImageNet-1K.}}
\label{tab:impl_pretraining}
\begin{tabular}{y{93}|x{98}}
\shline
config & ViT-S/16, ViT-B/16 \\
\shline
optimizer & AdamW \\
base learning rate & 1.5e-4 \\
weight decay & 0.05 \\
optimizer momentum & $\beta_1, \beta_2{=}$ 0.9, 0.95 \\
batch size & 4096 \\
learning rate schedule & cosine decay \\
pre-training epochs & 300, 300/800 \\
warmup epochs & 15, 15/40 \\
augmentation & random cropping  \ \ \&horizontal flip \\
perturbation budget &  2 \\
attack steps & standard:2, fast:1 \\
adapter & LayerNorm \\
adversarial distance loss & $L_2$ distance loss \\
loss ratio & 0.5 \\
adversarial ratio & standard:100\%, fast:25\% \\
pre-training resolution & 224 $\times$ 224 \\
\shline
\end{tabular}
\end{table}

\noindent \textbf{Fine-tuning on the ImageNet-1K} All the models are fine-tuned with image resolutions of 224 $\times$ 224. The Vit-S/16 and Vit-B/16 are further trained for 200/100 epochs, with 5 warmup epochs. The drop path rate~\cite{huang2016deep} is set to 0.1, and the batch size is 2048. For ViT-S/16, we set the base learning rate and the layer-wise learning rate decay to 1$e-$3 and 0.8. For ViT-B/16, the base learning rate and the layer-wise learning rate decay are set to 1$e-$3 and 0.7 for models pre-trained with 300 epochs, and set to 5$e-$4 and 0.65 for models pre-trained with 800 epochs. We train each model with strong data augmentation including label smoothing~\cite{szegedy2016rethinking}, mixup~\cite{zhang2018mixup}, cutmix~\cite{yun2019cutmix}, and randAugment~\cite{cubuk2020randaugment}. We use the global pooling feature rather than the class token during fine-tuning, following MAE. All models are fine-tuned using an image resolution of 224 $\times$ 224. And we also use the linear $lr$ scaling rule: $lr = base\_lr \times batch\_size /$256.
The default setting is shown in Table~\ref{tab:impl_finetuning}. 
And we also evaluate the robustness of models on ImageNet variants such as ImageNet-Real~\cite{beyer2020we}, ImageNet-V2~\cite{recht2019imagenet}, ImageNet-A~\cite{hendrycks2021natural}, ImageNet-R~\cite{hendrycks2021many}, ImageNet-Sketch~\cite{wang2019learning} and ImageNet-C~\cite{hendrycks2018benchmarking}.

\begin{table}[!hbtp]
\centering
\caption{{Fine-tuning setting on ImageNet-1K.}}
\label{tab:impl_finetuning}
\begin{tabular}{y{93}|x{98}}
\shline
config & ViT-S/16, ViT-B/16 \\
\shline
optimizer & AdamW \\
base learning rate & 1e-3, 1e-3/5e-4 \\
weight decay & 0.05 \\
optimizer momentum & $\beta_1, \beta_2{=}$ 0.9, 0.999 \\
layer-wise lr decay & 0.8, 0.7/0.65 \\
batch size & 2048 \\
learning rate schedule & cosine decay \\
training epochs & 200, 100 \\
warmup epochs & 5 \\
augmentation & RandAug (9, 0.5) \\
label smoothing & 0.1 \\
mixup & 0.8 \\
cutmix  & 1.0 \\
drop path rate & 0.1 \\
fine-tuning resolution & 224 $\times$ 224 \\
\shline
\end{tabular}
\end{table}

\noindent \textbf{Fine-tuning on the MS-COCO} 
The COCO dataset~\cite{lin2014microsoft} is commonly used as a benchmark for evaluating object detection frameworks. The Mask R-CNN framework~\cite{he2017mask} is adopted to predict bounding boxes and instance masks of objects for object detection and instance segmentation tasks. 
The pre-trained Vit-S/16 and Vit-B/16 are employed as the backbone for Mask R-CNN~\cite{he2017mask}. We first fine-tune Mask-RCNN using the COCO~\cite{lin2014microsoft} train2017 split and then evaluate its performance in terms of AP$^{box}$ and AP$^{mask}$ on the val2017 split. We employ multi-scale training where we resize the images to have a short side between 480 and 800 and a long side no greater than 1333. We use the AdamW optimizer with a learning rate of 2$e-$3/3$e-$3, a weight decay of 0.05, and a batch size of 16. The layer-wise decay rate is 0.75/0.65, and the drop path rate is 0.1/0.2 for ViT-S/16 and ViT-B/16. We adopt the 1$\times$ training schedule: 12 epochs with the learning rate decayed by 10$\times$ at epochs 9 and 11. We follow the implementation of MMDetection~\cite{chen2019mmdetection} for Mask R-CNN. More default setting is shown in Table~\ref{tab:impl_coco}.

\begin{table}[!hbtp]
\centering
\caption{{Fine-tuning setting on MS-COCO.}}
\label{tab:impl_coco}
\begin{tabular}{y{93}|x{98}}
\shline
config & ViT-S/16, ViT-B/16 \\
\shline
optimizer & AdamW \\
base learning rate & 2e-3,3e-3 \\
weight decay & 0.05 \\
optimizer momentum & $\beta_1, \beta_2{=}$ 0.9, 0.999 \\
layer-wise lr decay & 0.75, 0.65 \\
batch size & 16 \\
learning rate schedule & step decay \\
training epochs & 12 \\
warmup steps & 500 \\
drop path & 0.1, 0.2 \\
\shline
\end{tabular}
\end{table}

\begin{table*}[h]
\caption{Top-1 accuracy (\%) on ImageNet-1K and its IID variants. We use the MAE's 300-epoch training time as the reference for RPTT (Relative Pre-Training Time). * indicate that the results are reproduced with the official code.}\label{tab:1}
\centering
{
    \begin{tabular*}{\textwidth}{@{\extracolsep\fill}lccccc}
    \toprule
    Methods  & PT Epochs & Relative PT Time & IN & IN-Real & IN-V2 \\
    \midrule
    \multicolumn{5}{l}{\textit{Pre-training using ViT-S/16}} \\
    MoCo-v3 & 300 & - & {81.4} & - & - \\
    BEiT & 300 & 3.5$\times$ & 81.3 & - & - \\
    CIM-RESPIX & 300 & 6.3$\times$ & \textbf{81.5} & - & - \\
    MAE*  & 300 & 1$\times$  & 80.9 &86.9 &70.1 \\
    Fast AEMIM  & 300 & 1.3$\times$ & 81.2 & 87.0 & 70.5 \\
    AEMIM & 300 & 2.3$\times$ & {81.4} &\textbf{87.2} &\textbf{70.8} \\
    \midrule
    \multicolumn{5}{l}{\textit{Pre-training using ViT-B/16}} \\
    MoCo-v3 & 300 & - & 83.2 & - & - \\
    BEiT & 800 & 9.3$\times$ & 83.2 & - & - \\
    CIM-RESPIX & 300 & 6.3$\times$ & 83.3 & - & - \\
    CIM-RESPIX & 800 & 16.8$\times$ & 83.4 & - & - \\
    MAE*  & 300 & 1$\times$ & 82.9 &87.8 &72.1\\
    MAE*  & 800 & 2.7$\times$ & 83.3 & 87.8 & 72.7\\
    MAE\ \ \   & 1600 & 5.3$\times$ & 83.6 & 88.2 &73.2\\
    Fast AEMIM  & 300 & 1.3$\times$ & 83.3 & 87.8 & 72.9\\
    Fast AEMIM  & 800 & 3.5$\times$ & 83.6 & 88.1 & 73.5\\
    AEMIM  & 300 & 2.3$\times$ & 83.4 &87.9 &72.9 \\
    AEMIM  & 800 & 6.1$\times$ & \textbf{83.8} & \textbf{88.5} & \textbf{74.0} \\
    \bottomrule
    \end{tabular*}
}   
\end{table*}

\begin{table*}[tp]
\caption{Robustness evaluation on ImageNet OOD variants. The Top-1 accuracy (\%) metric is used for all variants except ImageNet-C, where mean corruption error (mCE) is used. The Score is calculated as the average of all results, and (1-mCE) is used for the calculation. * indicate that the results are reproduced with the official code.}
\label{tab:2} 
\centering  
{
    \begin{tabular*}{\textwidth}{@{\extracolsep\fill}lcccccc} 
    \toprule
    Methods  & PT Epochs & IN-A & IN-R & IN-S & IN-C $\downarrow$ & Average Score \\
    \midrule
    \multicolumn{6}{l}{\textit{Pre-training using ViT-S/16}} \\
    MAE*  & 300 & {24.5} & 45.5 & 32.0& {53.9} & 37.0 \\
    Fast AEMIM  & 300 & \textbf{24.6} & 45.9 & 32.2 & \textbf{53.5} & 37.3 \\
    AEMIM & 300 & 24.3 & \textbf{46.0} & \textbf{33.1} & {53.9}  & \textbf{37.4} \\
    \midrule
    \multicolumn{6}{l}{\textit{Pre-training using ViT-B/16}} \\
    MAE*  & 300 & 30.9 & 48.1 &34.2 &52.4 &40.2 \\
    MAE* & 800 & 35.3 & 48.3 & 34.4 & 51.9 & 41.5 \\
    MAE\ \ \  & 1600 & 35.9 & 48.3 & 34.5 & 51.7 & 41.8 \\
    Fast AEMIM  & 300 & 34.3 & 49.7 & 36.1 & 50.1 & 42.5 \\
    Fast AEMIM  & 800 & 37.6 & 49.7 &36.0 &49.6 & 43.4  \\
    AEMIM  & 300 & 34.5 & 49.6 & 35.6 &49.8 &42.5 \\
    AEMIM  & 800 & \textbf{38.0} & \textbf{50.6} & \textbf{36.5} & \textbf{49.3} & \textbf{44.0} \\
    \bottomrule
    \end{tabular*}  
}
\end{table*}

\begin{table*}[h]
\caption{Robustness results across different perturbation budgets $\epsilon$ under the FGSM attack for ViT-B/16. * indicate that the results are reproduced with the official code.}
\label{tab:robustness_fgsm}
\centering 
{
    \begin{tabular*}{0.9\textwidth}{@{\extracolsep\fill}lccccccc}
    \toprule
    \multirow{2}*{Methods} & \multirow{2}*{PT Epochs} & \multicolumn{6}{c}{Accuracy under FGSM} \\ \cmidrule{3-8}
    & & $\epsilon=0$& $\epsilon=1$& $\epsilon=2$& $\epsilon=4$& $\epsilon=8$& Average Acc. \\
    \midrule
    MAE* & 800 & 83.3 & 42.6& 37.3 & 33.9& 31.4& 45.7\\
    MAE & 1600 & 83.6 & 42.8 & 38.4 & 35.2& \textbf{33.0}& 46.6\\
    Fast AEMIM & 800 & 83.6 & 44.0 &38.3 & 33.3& 30.8&46.0\\
    AEMIM & 800 & \textbf{83.8} &\textbf{44.3} &\textbf{38.7}&\textbf{35.3}&32.1& \textbf{46.9}\\
    \bottomrule
    \end{tabular*}  
}
\end{table*}

\begin{figure*}[t]
\begin{center}
\includegraphics[width=0.65\linewidth]{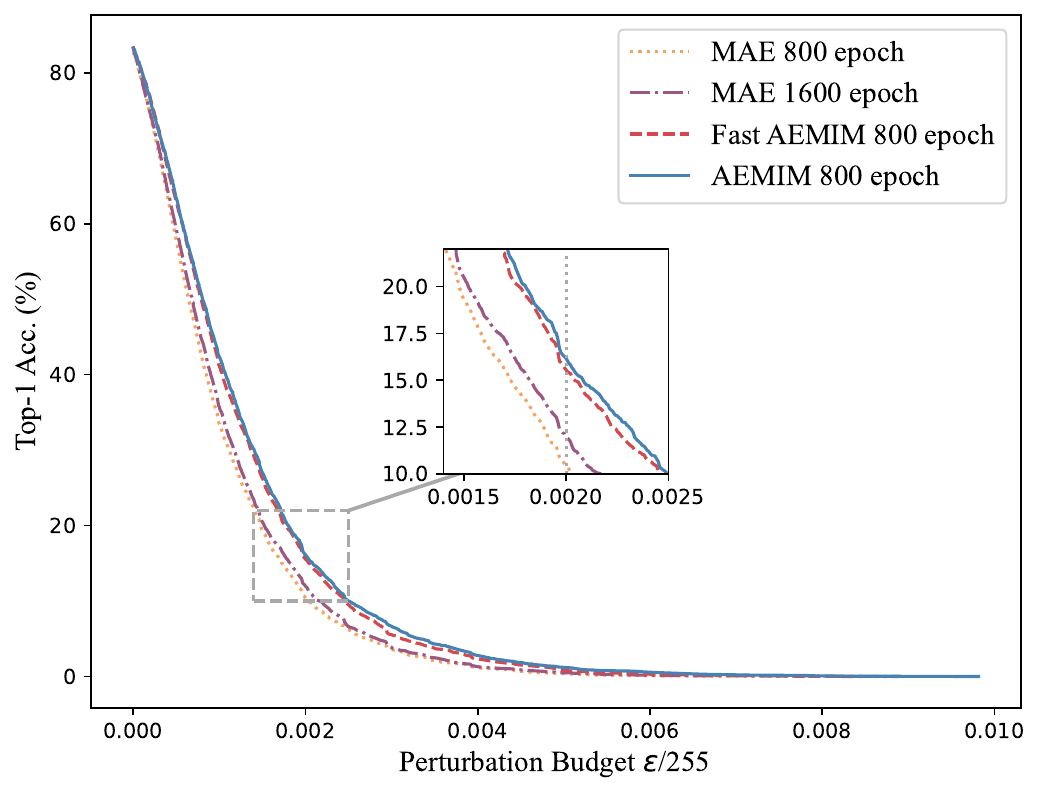}
\end{center}
\caption{Robustness curves of Top-1 accuracy (\%) vs. perturbation budget under the PGD attack for ViT-B/16. }
\label{fig:robustness_pgd}
\end{figure*}

\subsection{ImageNet and its IID variants Results}
As mentioned above, the encoders are fine-tuned on the ImageNet-1K dataset in a supervised manner after the pre-training stage. 
Since we adopt MAE as the basic MIM framework without altering any MIM strategies, we mainly compare our method with MAE in experiments to demonstrate its gains. It should be emphasized that our method is independent of any specific MIM methods or model structures, and can also be used as a plugin to enhance other basic MIM methods. 

Table~\ref{tab:1} shows the results of Vit-S/16 and Vit-B/16 on the ImageNet validation dataset and its IID variants ImageNet-Real, ImageNet-V2. The Vit-S/16 pre-trained by AEMIM for 300 epochs reaches a fine-tuning accuracy of 81.4\%, which has an improvement of 0.5\% compared with MAE. For Vit-B/16, our 300-epoch model achieves 83.4\%, which is even slightly higher than the model of MAE at 800 epochs. When enlarging the training scheduler to 800 epochs, our model of Vit-B/16 achieves 83.8\% Top-1 accuracy, which is 0.5\%/0.2\% higher than the MAE baseline at 800/1600 epochs. For the ImageNet IID variants, our method also outperforms MAE under different training schedulers. For example, our best model of Vit-B/16 outperforms MAE by 0.3\% and 0.8\% on ImageNet-Real and ImageNet-V2, respectively. The results show that our Fast AEMIM outperforms the baseline MAE and CIM, requiring only $1.3\times$ the training cost of MAE for each epoch, which is much faster than the previous CIM. The model trained with Fast AEMIM for 800 epochs exhibits improved generalization compared to the model trained with MAE for 1600 epochs while utilizing only about $65\%$ of the training cost. The detailed training cost analysis is shown in Sec.~\ref{sec:4.6}.

\subsection{ImageNet OOD Variants Results}
We test the robustness of our models on ImageNet OOD variants, including  ImageNet-A, ImageNet-R, ImageNet-Sketch, and ImageNet-C. The Top-1 accuracy is used as the evaluation metric for all datasets, but for ImageNet-C. For ImageNet-C, we use mean corruption error (mCE) as the metric, and a lower value is better. 
And (1$-$mCE) is used to get the final Score, which is the average of all the results.

Table~\ref{tab:2} compares the robustness of our models to the MAE baseline. The results show that our Vit-B/16 model trained with AEMIM for only 300 epochs outperforms MAE's best model trained for 1600 epochs. Specifically, our 300/800-epoch Vit-B/16 model trained with AEMIM achieves an average score increase of 0.7/2.2\% compared to the best model of MAE. The model trained with Fast AEMIM also shows a 0.7/1.6\% improvement over MAE's best model. The results demonstrate that our method assists the model in learning more generalized representations, thereby significantly enhancing the model's generalization to OOD datasets.

\subsection{Adversarial Robustness Results}
In addition to evaluating the robustness of our method on OOD datasets, we also evaluate its adversarial robustness. During the pre-training stage, we utilize the PGD attack with 1 or 2 attack steps to generate adversarial examples, which is essentially closer to the FGSM attack. Thus, we first employ FGSM as the attacker. In the evaluation, we employ the same set of 5000 images as defined in the adversarial robustness benchmark ARES-Bench~\cite{liu2023comprehensive}, which are drawn from the ImageNet validation dataset. Table~\ref{tab:robustness_fgsm} presents the accuracy of different models across varying perturbation budge $\epsilon$. The results suggest that our method achieves a slight improvement in adversarial robustness compared to MAE under the FGSM attack. Our method performs better at small $\epsilon$, while demonstrating slightly weaker performance at larger $\epsilon$.

Then, we escalate the attack to the stronger PGD with 20 attack steps. During the fine-tuning stage, we conduct standard training rather than adversarial training. Consequently, when confronted with stronger attacks, a substantial increase in $\epsilon$ can quickly lead to a sharp drop in accuracy to 0. To address this issue, we draw the robustness curves of classification accuracy vs. perturbation budget based on ARES-Bench, as illustrated in Figure~\ref{fig:robustness_pgd}. The results indicate that our method exhibits significantly improved adversarial robustness compared to MAE when facing stronger attacks, although it still falls short of models trained with adversarial training. As an example, when the perturbation budget $\epsilon/255$ is 0.002, the 800-epoch model trained using AEMIM exhibits an accuracy improvement of (16.0$-$11.9)\%=4.1\% compared to the 1600-epoch model trained using MAE.

\subsection{Downstream Task Results}
For the downstream task, we transfer our pre-trained models to object detection and instance segmentation tasks on the COCO dataset. We report the AP$^{box}$ and AP$^{mask}$ of our method for object detection and instance segmentation, comparing with the baseline in Table~\ref{tab:3}. The results of the baseline are reproduced using the official code under our experimental settings. For Vit-S/16, our Fast AEMIM/AEMIM surpass MAE by 0.4/1.6\% and 0.4/1.4\% for AP$^{box}$ and AP$^{mask}$. For Vit-B/16, Fast AEMIM/AEMIM improve AP$^{box}$ and AP$^{mask}$ by 0.8/0.4\% and 0.4/0.5\%, compared with MAE at 800/1600 epoch. The results demonstrate that our method is also capable of facilitating the model to learn better representations for downstream vision tasks.
\begin{table}[h]
\caption{COCO results for object detection and instance segmentation in terms of AP$^{box}$ (\%) and AP$^{mask}$ (\%). All methods are based on the Mask R-CNN. * indicate that the results are reproduced with the official code.}
\label{tab:3}  
\centering  
{
    \begin{tabular*}{\linewidth}{@{\extracolsep\fill}lccc} 
    \toprule
    Methods & PT Epochs & AP$^{box}$ &AP$^{mask}$ \\
    \midrule
    \multicolumn{4}{l}{\textit{Pre-training using ViT-S/16}} \\
    BEiT & 300 & 35.6 &32.6 \\
    MAE* & 300 & 38.8 &35.1 \\
    Fast AEMIM & 300 & 39.2 & 35.5 \\
    AEMIM & 300 & \textbf{40.4} & \textbf{36.5} \\
    \midrule
    \multicolumn{4}{l}{\textit{Pre-training using ViT-B/16}} \\
    BEiT & 800 & 35.6 & 32.6 \\
    MAE* & 300 & 45.8 & 40.7 \\
    MAE* & 800 & 47.2 & 42.0  \\
    MAE* & 1600 & 48.3 &42.5 \\
    Fast AEMIM & 800 & 48.0 &42.4  \\
    AEMIM & 300 & 46.4 &41.3  \\
    AEMIM & 800 & \textbf{48.7} & \textbf{43.0}  \\
    \bottomrule
    \end{tabular*}  
}
\end{table}

\subsection{Pre-training Cost Analysis}
\label{sec:4.6}
We compare the pre-training cost of our method with the baseline methods BEiT~\cite{bao2021beit}, MAE~\cite{he2022masked} and CIM~\cite{fang2023corrupted}. In the CIM framework, aside from the standard auto-encoder model, there are additional components, including the BEiT-Style generator and the DALL-E tokenizer. Within this setup, the auto-encoder and generator are actively being trained, while the tokenizer remains in a frozen state. This introduces a significant additional computational burden, causing CIM to be even slower than BEiT. The pre-training time for each epoch of CIM is approximately 1.8$\times$ that of BEiT. Compared to CIM, our method incurs additional computational burden due to the increased gradient backpropagation needed for generating adversarial examples, as well as the co-training of generated adversarial data. In this paper, our method is founded on the MAE framework. Thus, the pre-training time per epoch for AEMIM is about 2.2-2.3$\times$ that of MAE. In the case of Fast AEMIM, we considerably decrease the quality and quantity of generated adversarial examples, leading to a significant reduction in the additional computational cost. The pre-training time per epoch for Fast AEMIM is only around 1.3$\times$ that of MAE. 

Table~\ref{tab:1} and Table~\ref{tab:2} presents the relative pre-training time of different methods and their corresponding performance. For Vit-S/16, although CIM slightly outperforms our method on ImageNet validation, its computational cost far exceeds ours. When compared to MAE, our method consistently shows notable improvements across different datasets. Regarding Vit-B/16, firstly, the 300-epoch models trained by AEMIM and Fast AEMIM exhibit competitive results with CIM, while incurring considerably lower computational costs. In comparison to MAE, the 300-epoch model trained by Fast AEMIM achieves comparable results to the 800-epoch model trained by MAE, and even demonstrates better performance on ImageNet OOD variants, while utilizing only 49\% of the computational cost. Similarly, the 800-epoch model trained with Fast AEMIM attains superior results compared to the 1600-epoch model trained by MAE, consuming only 65\% of the computational cost. Furthermore, the 800-epoch model trained with AEMIM, despite incurring 1.1-1.2$\times$ the computational cost of the 1600-epoch MAE model, demonstrates significantly enhanced generalization and robustness. Overall, these results indicate that our method offers a superior performance-to-efficiency ratio compared to MAE and CIM.

\subsection{Ablations}
\begin{table*}
  \centering
  \caption{{AEMIM ablation experiments} with Vit-S/16 on ImageNet-1K. We report fine-tuning (ft) Top-1 accuracy (\%). The selected settings are \underline{underlined}.}
  \label{tab:ablations}
  \begin{tabular}{cc}
  \begin{subtable}{0.45\linewidth}
    \centering
    \setlength{\tabcolsep}{18pt}
    {
    \captionsetup{justification=centering}
    \caption{\textbf{Perturbation budget}. A moderate perturbation budget ($\epsilon$=2) works better for the regularization of pre-training.}
    \label{tab:attack_epsilon}
    \begin{tabular}{cc}
      \toprule
      $\epsilon$ & Top-1 \\
      \midrule
      1 & 81.3 \\
      \underline{2} & {\textbf{81.4}} \\
      4 & 81.2 \\
      \bottomrule
    \end{tabular}
    }
  \end{subtable}
  &
  \begin{subtable}{0.45\linewidth}
    \centering
    \setlength{\tabcolsep}{16pt}
    {
    \captionsetup{justification=centering}
    \caption{\textbf{Attack steps}. The appropriate number of attack steps ($T$=2) ensures both the training efficiency and accuracy.}
    \label{tab:attack_iteration_num}
    \begin{tabular}{cc}
      \toprule
      $T$ & Top-1 \\
      \midrule
      1 & 81.2 \\
      \underline{2} & {\textbf{81.4}} \\
      3 & \textbf{81.4} \\
      \bottomrule
    \end{tabular}
    }
    \end{subtable}
    \\
    \begin{subtable}{0.45\linewidth}
    \centering
    \setlength{\tabcolsep}{10pt}
    {
    \captionsetup{justification=centering}
    \caption{\textbf{Adapter}. Both the classification token and the normalization layer work well as the adapter.}
    \label{tab:adapter}
    \begin{tabular}{cc}
      \toprule
      Adapter & Top-1 \\
      \midrule
      w/o & 81.2 \\
      Cls token & {\textbf{81.4}} \\
      \underline{LN} & \textbf{81.4} \\
      \bottomrule
    \end{tabular}
    }
  \end{subtable}
  &
  \begin{subtable}{0.45\linewidth}
    \centering
    \setlength{\tabcolsep}{10pt}
    {
    \captionsetup{justification=centering}
    \caption{\textbf{Adversarial loss}. $L_2$ distance loss performs best to generate compatible adversarial examples.}
    \label{tab:adv_loss}
    \begin{tabular}{cc}
      \toprule
      Adv loss & Top-1 \\
      \midrule
      Rec loss & 81.1  \\
      KLD & 81.3  \\
      \underline{$L_2$ dist} & {\textbf{81.4}}  \\
      \bottomrule
    \end{tabular}
    }
  \end{subtable}
  \end{tabular}
\end{table*}

In this section, we first conduct a series of ablation experiments to investigate the impact of important components and validate design choices in our method. Here we focus on studying how adversarial examples affect MIM pre-training instead of the MIM strategies. The corresponding ablation studies are performed using a Vit-S/16 model that is pre-trained for 300 epochs with AEMIM, followed by 200 epochs of fine-tuning on ImageNet-1K. We adopt the normalization layer as the adapter and $L_2$ distance loss as the adversarial loss by default. And we set $\epsilon=2, T=2, \lambda=0.5$ for perturbation budget, number of attack steps, and loss ratio, unless specified. 

\noindent\textbf{Perturbation budget.} The perturbation budget affects the regularization strength of adversarial examples. Table~\ref{tab:attack_epsilon} presents the results of using different perturbation budgets to generate adversarial examples during pre-training. The results show that a moderate perturbation budget ($\epsilon$ = $2$) is more effective for regularizing pre-training.

\noindent\textbf{Attack steps.} The number of attack steps has an impact on the quality of generated adversarial examples. As the number of steps increases, the quality of adversarial examples improves, but the efficiency of generation decreases. 

The results in Table~\ref{tab:attack_iteration_num} show that generating adversarial examples with two steps is sufficient. We finally set the number of steps to 2 for efficiency.

\begin{table*}[h]
\caption{Ablation experiment for MIM framework on ImageNet-1K.  We utilize SimMIM as the basic MIM framework and boost its performance with our approach. And we select Swin-B as the backbone model.}
\label{tab:other_mim} 
\centering  
{
    \begin{tabular*}{0.9\textwidth}{@{\extracolsep\fill}lcccccccc}
    \toprule
    Methods  & PT Epochs & IN & IN-Real & IN-V2 & IN-A & IN-R & IN-S & IN-C $\downarrow$  \\
    \midrule
    \multicolumn{6}{l}{\textit{Pre-training using Swin-B}} \\
    SimMIM  & 100 &83.5 & 88.2&\textbf{72.8} & 34.9 & 47.5 & 34.5 & 53.4 \\
    AEMIM  & 100 &\textbf{83.7} &\textbf{88.3} &72.7 & \textbf{36.5} & \textbf{48.5} & \textbf{35.8} & \textbf{52.4}  \\
    \bottomrule
    \end{tabular*}  
}
\end{table*}

\noindent\textbf{Adapter.} Adapters are crucial for co-training with data from different domains. Table~\ref{tab:adapter} shows the impact of adapters on our method. Unlike in supervised learning, our approach still yields some improvements in pre-training even when the adapters are removed. For the type of adapter, the results demonstrate that both the classification token and normalization layer work well with our method.

\noindent\textbf{Adversarial loss.} Table~\ref{tab:adv_loss} shows the results of using different adversarial loss to generate adversarial examples during pre-training. 
As Figure~\ref{fig:2} indicates, using the reconstruction loss as the adversarial loss directly can shift the pre-training focus towards the auxiliary pretext task, thus impacting the model's representation learning. Nevertheless, using the reconstruction loss can still be advantageous for pre-training, highlighting that our designed auxiliary pretext task itself can also help the model learn good representations. As for the type of adversarial loss, $L_2$ distance loss works best in our method.

\begin{figure}[tp]
\begin{center}
\includegraphics[width=\linewidth]{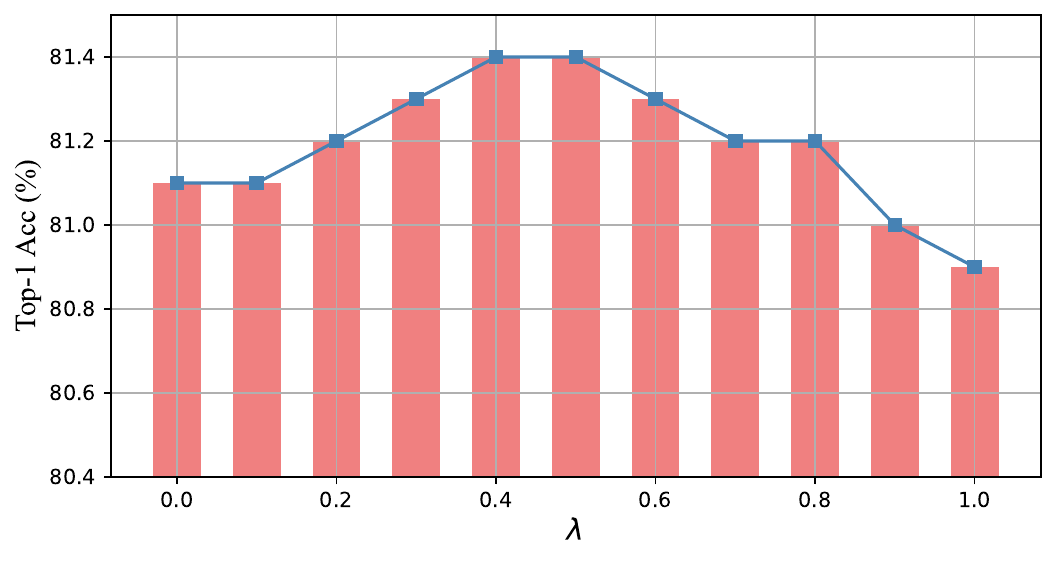}
\end{center}
\caption{Pre-training with different loss ratio $\lambda$. We report the Top-1 accuracy (\%) on ImageNet-1K.}
\label{fig:loss_ratio}
\end{figure}

\noindent\textbf{Loss ratio.} The loss ratio balances the two pretext tasks during pre-training. Figure~\ref{fig:loss_ratio} illustrates the results with different loss ratios $\lambda$. And we empirically set $\lambda=0.5$ in our method.

We then conduct a light ablation experiment to verify that the proposed method can boost other MIM frameworks and architectures. In the experiment, we employ SimMIM~\cite{xie2022simmim} as the basic MIM framework and select Swin-B~\cite{liu2021swin} as the backbone model. We follow all of the MIM training settings from SimMIM. For the attack settings, we adopt the default settings from AEMIM. That means we adopt the normalization layer as the adapter and $L_2$ distance loss as the adversarial loss by default. And we set $\epsilon = 2$, $\lambda = 0.5$, $\alpha=100\%$ for perturbation budget, loss ratio, and adversarial ratio. Here we set the number of attack steps as $T = 1$ for efficiency. Table~\ref{tab:other_mim} presents the performance of models pre-trained for 100 epochs. The results demonstrate that our method achieves better performance compared to SimMIM, especially on OOD datasets. On ImageNet OOD variants, our method achieves an average improvement of 1.2\% compared to SimMIM. This further validates that our method can indeed enhance other MIM frameworks and model architectures.

\section{Discussion}
Our method serves to assist the current MIM methods in approaching their performance limits, rather than directly improving those limits, which would necessitate the design of improved MIM strategies and model architectures. 
Additionally, in our framework, we primarily use traditional noise-based adversarial examples. However, the community has recently seen the emergence of new types of adversarial examples, such as on-manifold adversarial examples~\cite{Stutz_2019_CVPR,xiao2022understanding,Bar_Tal_2023_ICCV,kazemi2023minimally}. These novel examples maintain their adversarial properties while exhibiting better image quality, closer to the original data distribution. As a result, they perform better in enhancing the performance of visual models~\cite{liu2023improving}. 
Using improved adversarial examples in our framework could enhance performance further, but there is a potential risk of reduced training efficiency due to the slower generation of these examples. We leave it to the future work.

\section{Conclusion}
In this paper, we propose a novel method called AEMIM, which incorporates adversarial examples into the masked image modeling (MIM) technique for pre-training. Our approach introduces a new auxiliary pretext task of reconstructing the corresponding adversarial examples, in addition to the original pretext task of reconstructing masked normal images. We also design a novel adversarial attack to synthesize adversarial examples for MIM pre-training, using the distance between the encoder's feature representation of the adversarial data and the clean data as the adversarial loss. To enhance efficiency, we have also developed a faster version of our method. This optimized version maintains the core functionalities while significantly reducing computation time, making it suitable for applications where speed is crucial. Our method surpasses the baseline on various datasets, including ImageNet-1K, ImageNet variants, and the downstream COCO datasets, demonstrating its effectiveness in enhancing the current MIM method. We believe that this work can provide the community with a fresh perspective on the relationship between adversarial examples and MIM approaches, potentially serving as inspiration for future research endeavors.

\backmatter





\bibliography{sn-bibliography}

\end{document}